\documentclass[twoside,11pt]{article}

\usepackage{automl2020}

\usepackage{times}
\usepackage{soul}
\usepackage{url}
\usepackage[utf8]{inputenc}
\usepackage[small]{caption}
\usepackage{graphicx}

\usepackage{amsmath}
\usepackage{amsthm}
\usepackage{booktabs}
\usepackage{algorithm}
\usepackage{textalpha}
\usepackage[noend]{algpseudocode}
\usepackage{tikz}
\usetikzlibrary{arrows, shapes}
\usepackage{amssymb}
\usepackage{subfig}
\usepackage{afterpage}
\usepackage{multirow}
\usepackage{float}
\usepackage{pbox}
\usepackage{array}
\usepackage{svg}
\usepackage[symbol]{footmisc}

\jmlrheading{M. Gupta, S. Aravindan, A. Kalisz, V. Chandrasekhar and L. Jie}

\ShortHeadings{Learning to Prune Deep Neural Networks via Reinforcement Learning}{Gupta, Aravindan, Kalisz, Chandrasekhar, Jie}
\firstpageno{1}

\begin{document}

\title{Learning to Prune Deep Neural Networks\\via Reinforcement Learning}

\author{
        \name Manas Gupta
        \email Manas\_Gupta@i2r.a-star.edu.sg \\
        \addr I2R, ASTAR Singapore
        \AND 
        \name Siddharth Aravindan  
        \email siddharth.aravindan@comp.nus.edu.sg   \\
        \addr National University of Singapore
        \AND 
        \name Aleksandra Kalisz\footnotemark[2]  
        \email kalisz.ola@gmail.com  \\
        \addr I2R, ASTAR Singapore
        \AND
        \name Vijay Chandrasekhar\footnotemark[2]  
        \email vijay@i2r.a-star.edu.sg  \\
        \addr  I2R, ASTAR Singapore
        \AND
        \name Lin Jie  
        \email lin-j@i2r.a-star.edu.sg \\
        \addr I2R, ASTAR Singapore
    }
\footnotetext[2]{For contributions made in 2019}

\maketitle

\begin{abstract}%<- trailing '%' for backward compatibility of .sty file
This paper proposes PuRL - a deep reinforcement learning (RL) based algorithm for pruning neural networks. Unlike current RL based model compression approaches where feedback is given only at the end of each episode to the agent, PuRL provides rewards at every pruning step. This enables PuRL to achieve sparsity and accuracy comparable to current state-of-the-art methods, while having a much shorter training cycle. PuRL achieves more than 80\% sparsity on the ResNet-50 model while retaining a Top-1 accuracy of 75.37\% on the ImageNet dataset. Through our experiments we show that PuRL is also able to sparsify already efficient architectures like MobileNet-V2. In addition to performance characterisation experiments, we also provide a discussion and analysis of the various RL design choices that went into the tuning of the Markov Decision Process underlying PuRL. Lastly, we point out that PuRL is simple to use and can be easily adapted for various architectures.
\end{abstract}

\section{Introduction}

Neural network efficiency is important for specific applications, e.g., deployment on edge devices and climate considerations~\cite{strubell2019energy}. Weight pruning has emerged as a viable solution methodology for model compression~\cite{Han2015DeepCC}, but pruning weights effectively remains a difficult task --- the search space of pruning actions is large, and over-pruning weights (or pruning them the wrong way) leads to deficient models~\cite{frankle2019stabilizing,imagenet_cvpr09}.

In this work, we approach the pruning problem from a decision-making perspective, and propose to automate the weight pruning process via reinforcement learning (RL). RL provides a principled and structured framework for network pruning, yet has been under-explored. There appears to be only one existing RL-based pruning method, namely AutoML for Model Compression (AMC) \cite{he2018amc}. Here, we build upon AMC and contribute an improved framework: Pruning using Reinforcement Learning (PuRL). 

Compared to AMC, PuRL rests on a different  Markov Decision Process (MDP) for pruning. One key aspect of our model is the provision of ``dense rewards'' --- rather than rely on the ``sparse'' rewards (given only at the end of each episode), we shape the reward function to provide reward feedback at each step of the pruning process. This results in a far shorter training cycle and decreases the number of training episodes required by as much as 85\%. The remaining design changes are informed by ablation-style experiments; we discuss these changes in detail and elucidate the trade-offs of different MDP configurations.

\section{Related Work}
\label{RelatedWork}

Various techniques have been proposed for compressing neural networks \cite{cheng2017survey}. Pruning comes out as a general approach not having restrictions in terms of the tasks it is applicable to. However, pruning too has a large search space size and hence, traditionally, human expertise has been relied upon to do pruning. But with the advent of new search techniques like deep reinforcement learning, we can now automate the process of pruning. In Runtime Neural Pruning, \cite{NIPS2017_6813} demonstrate one early approach for using RL to do pruning. They use RL to do a sub-network selection during inference. Thus, they actually don't really prune the network, but select a sub-network to do inference. \cite{he2018amc} demonstrate the first use of RL for pruning. However, they only reward the agent at the end of an episode (sparse rewards) and don't give it any reinforcement at each step in the episode. This slows down the learning process of the RL agent. 

We improve upon this by creating a novel training procedure that rewards the agent at each step of the episode (dense rewards) and achieves faster convergence. Our approach is also general in nature and can be easily adapted for different architectures. We compare and report our performance with regards to AMC and other state-of-the-art pruning algorithms on the ImageNet dataset.

\section{Pruning using Reinforcement Learning (PuRL)}
\label{Approach}

This section details PuRL, our reinforcement learning method for network pruning. We formalize network pruning as an MDP; we specify the constituent elements, along with intuitions underlying their design. 

\subsection{Pruning as a Markov Decision Problem}

We model the task of pruning a neural network as a Markov Decision Problem (MDP). We formulate and construct each element of the MDP tuple i.e.  $\langle\mathcal{S},\mathcal{A},\mathcal{R},\mathcal{T},\gamma \rangle$ to enable us to use RL for pruning. In the below subsections, we elaborate on each of the tuple elements.  

\subsubsection{State Representation} 

We represent the network state $s$ through a tuple of features. We experiment with two different kind of representations. The first is a simple representation scheme consisting of three features, $s = \langle l, a, p \rangle$, where $l$ is the index of the layer being pruned, $a$ is the current accuracy achieved on the test set (after retraining) and $p$ corresponds to the proportion of weights pruned thus far. The attributes serve as indicators of the network state. The second representation is a higher dimensional representation aimed at capturing more granular information on the state of the network. It is formulated as, $s = \langle a_1, p_1, a_2, p_2, .., a_n, p_n \rangle$, where $a_i$ is the test accuracy after pruning layer $i$ and doing retraining and $p_i$ is the sparsification percentage of layer $i$. $s$ is a tuple of zeros at the start of every episode. Each tuple element is updated progressively as layer $i$ is pruned. We report results from both state representations in the ablation experiments.

\subsubsection{Action Space}
\label{sec:expt_res}

The action space consists of actions where each action corresponds to an \textalpha\ value which decides the amount of pruning. We use a magnitude threshold derived from the standard deviation of the weights of a layer as our pruning criteria. We prune all weights smaller than this threshold in absolute magnitude. The set of weights that get pruned when an $\textrm{Action}_{i}(\alpha)$ is taken for layer $i$ are given by Equation \ref{eq:prune_set}. 
\begin{equation}
\textrm{Weights Pruned}_i(\alpha) = \{w \mid \mathopen |w \mathclose | < \alpha \sigma(w_i) \}
\label{eq:prune_set}
\end{equation}
where $\sigma(w_i)$ is the standard deviation of weights in layer \textit{i}. To further reduce the search complexity, we also experiment with increasing the step size of our actions from 0.1 to 0.2, to sample less actions but still achieve same target pruning rates. We report results in the ablation experiments. 
We use the same action space for all layers in the network i.e. $\alpha \in \{0.0, 0.1, 0.2, \cdots, 2.2\}$. This is in contrast to current approaches like AMC and State of Sparsity which set a different pruning range for initial layers, in order to prune them by a lesser amount. Our approach is hence more general in this aspect.

\subsubsection{Reward Function}
Since, the RL agent learns the optimal sparsification policy based on the objective of maximization of total reward per episode, reward shaping helps in faster convergence \cite{ng1999policy}. We formulate the total reward to be an accumulation of sub-rewards depending upon test accuracy and sparsification achieved. The reward function corresponding to a state $s$ is given in Equation \ref{eq:prune_reward}. Here, $A(s)$ and $P(s)$ denote the test accuracy and sparsity at state $s$, $T_A$ and $T_{P}$ denote the desired accuracy and sparsity target set by the user, and $\beta$ corresponds to a fixed scaling factor of 5. 
\begin{equation}
\textrm{R(s)} = - \beta ( \max (1 - A(s)/T_{A} ,  0) + \max (1 - P(s)/T_{P} , 0)) 
\label{eq:prune_reward}
\end{equation}
The reward design ensures that the agent jointly optimises for the desired sparsity and accuracy.

\subsection{The PuRL Algorithm}

We design the PuRL algorithm, to be fast and efficient, when solving the above-mentioned MDP. The first aspect of this is the choice of a good RL agent. The second and more important consideration is the rewards scheme i.e. sparse vs. dense rewards. We elaborate on each of these in the below subsections.

\subsubsection{Choice of RL Agent}

To solve the MDP we choose amongst various available RL algorithms. Our primary focus is on sample efficiency and accuracy. Deep Q-Network (DQN) \cite{mnih2013playing}, a form of Q-learning, does a very fast exploration, however, it is not very stable. Through careful design of our reward structure, we make DQN stable and hence, utilise it for doing pruning.

\subsubsection{Making RL Fast: Dense Rewards}

The pruning procedure consists of pruning the weights in a layer based on their magnitude first. The remaining weights are then retrained to get back the accuracy. Retraining is an important aspect of this process. 

By setting a different \textalpha\ for each layer, we try to prune away the maximum redundancy specific to each layer. As mentioned in section \ref{RelatedWork}, one way this has been done is to 1) assign alphas to each layer 2) prune each layer and 3) retrain the pruned network in the end. While this method works, as shown by \cite{he2018amc}, it may not be the fastest since it does not directly ascribe accuracy to the \textalpha\ of each layer. In other words, since retraining is only conducted after pruning all the layers and not after each layer (sparse rewards), the network cannot directly infer how accuracy is linked to \textalpha\ of each layer. This might elongate the training period since more samples are required to deduce effect of \textalpha\ of each layer to the network's final accuracy. 

\begin{figure}[t]
\centering
\includegraphics[width=10cm]{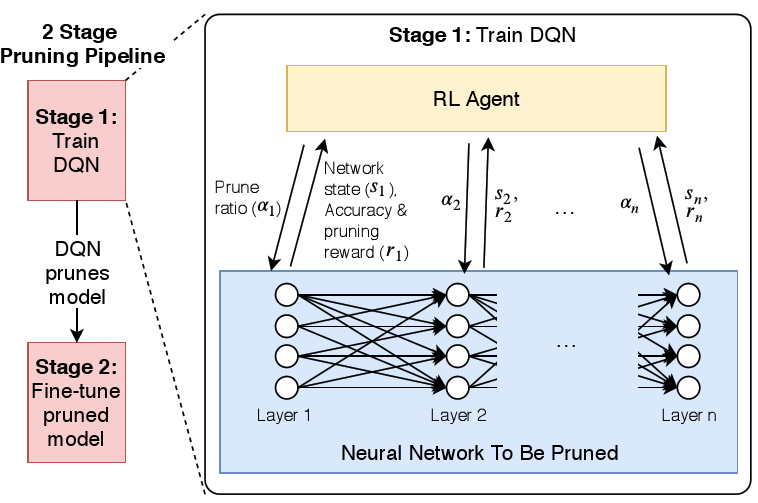}
\caption{A high level view of the PuRL algorithm with dense rewards. PuRL assigns a unique compression ratio \textalpha\ to each layer. It then gets feedback on the test accuracy and sparsity achieved, after pruning that layer. This is in contrast to current approaches which only give feedback at the end of pruning the whole network. As a result, PuRL learns the optimal sparsity policy 85\% faster than current approaches}
\label{fig:rl_autoprune}
\end{figure}

We try to mitigate this by giving rewards after pruning each layer as opposed to giving them at the end of the episode (dense rewards). We retrain the network after pruning each layer to get a test accuracy value. We do retraining by using only a small training set of 1000 images in the case of ImageNet experiments, so as not to add training overhead. We measure accuracy after each layer is pruned and pass it to the agent through the reward and state embedding. As mentioned in section \ref{Experiments}, this method of giving dense rewards helps achieve convergence much faster compared to giving sparse rewards. A high level view of the PuRL algorithm is presented in Figure \ref{fig:rl_autoprune}. 

\section{Experiments \& Analysis}
\label{Experiments}

In this section, we describe computational experiments comparing PuRL to ablated variants, as well as baseline and state-of-the-art methods. Our primary goal was to clarify the effect of different design choices (described in section \ref{ablation}) to the pruning performance. Secondly, we demonstrate that PuRL achieves comparable results to state-of-the-art while using a 85\% shorter RL training cycle by testing it on CIFAR-100 and ImageNet datasets and different architectures like ResNet-50, MobileNet-V2 and WideResNet-28-10 (refer to section \ref{imagenet} and \ref{imagenet_app}). Lastly, we showcase the generalization ability of PuRL by using the exact same settings to prune all architectures on ImageNet. 

\subsection{Understanding the RL Design Space}
\label{ablation}

We conduct a series of ablation experiments to understand what components of the RL design space help make a good RL agent. The choices that give superior result over the baseline are then eventually used. Due to space constraints, we elaborate on some choices in Appendix \ref{ablation_app}. We experiment on the ResNet-50 architecture trained on the CIFAR-10 dataset. We set a target sparsity of 60\% and target accuracy of 95\% for our agent (via the reward function), in all experiments.

\begin{table*}[!ht]
\small
\begin{tabular}{m{32mm}m{6mm}m{8mm}m{8mm}m{8mm}m{10mm}m{10mm}m{16mm}m{16mm}}
\toprule
\multirow{2}{*} {Experiment} &  \multirow{2}{*} {State} &  \multirow{2}{*} {Action} & \multicolumn{4}{c} {Reward Space} &  \multirow{2}{*}{Accuracy\%} &  \multirow{2}{*}{Sparsity\%}  \\\cmidrule(lr){4-7}
&Space&Size& \centering Prune Penalty& \centering Acc. Penalty& \centering Acc. Upside&\centering Cubic Upside&& \\%\hline
\midrule
Sparse Rewards & \centering3 & \centering0.1 & \centering\checkmark & \centering\checkmark  & & &68.0$\pm$13.9&66.1$\pm$12.7\\ %\hline
Dense Rewards (Baseline) & \centering3 & \centering0.1 & \centering\checkmark & \centering\checkmark  & & &91.8$\pm$2.0&70.1$\pm$2.3\\ %\hline
Baseline + Magnitude Target & \centering3 & \centering0.1 & \centering\checkmark & \centering\checkmark & & &86.6$\pm$3.9&60.3$\pm$0.8\\ %\hline
Baseline + Reward 2 & \centering3 & \centering0.1 & \centering \checkmark & \centering\checkmark & \centering\checkmark & &91.4$\pm$0.7&68.3$\pm$2.2\\ %\hline
Baseline + Reward 3 & \centering3 & \centering0.1 & \centering  \checkmark& \centering \checkmark& \centering \checkmark & \centering \checkmark &90.7$\pm$0.2&69.4$\pm$2.9\\ %\hline
Baseline + Action 2 & \centering3 & \centering0.2 & \centering\checkmark & \centering\checkmark  & & &92.2$\pm$0.5&70.6$\pm$0.6\\ %\hline
Baseline + State 2 & \centering108 & \centering0.1 & \centering\checkmark & \centering\checkmark  & & &78.9$\pm$8.6&72.2$\pm$4.5\\ %\hline
\bottomrule
\end{tabular}
\caption{Ablation results on perturbing State, Action and Reward spaces for the PuRL algorithm on the CIFAR-10 dataset. Error denotes standard error as measured on 3 trials. Dense rewards outperform sparse rewards by a huge margin on accuracy (rows 1 \& 2). Stepping the action space by 0.2 (row 6) leads to a Pareto dominant solution over the baseline (row 2)}
\label{tab:ablation_results}
\end{table*}

\subsubsection{Are Dense Rewards better than Sparse Rewards?}

We compare sparse rewards i.e. rewards given to the agent only at the end of the episode and dense rewards i.e. rewards at each step of the episode, and try to answer which is better. Referring to Table \ref{tab:ablation_results}, we compare sparse rewards (row 1) to dense rewards (row 2). Our dense rewards approach outperforms the sparse rewards by a huge margin, 4\% on sparsity and 24\% on accuracy. Dense rewards help the agent learn much faster by guiding the agent at each step instead of only at the end of the episode. We then use dense rewards as our baseline to conduct all further ablations.

\subsubsection{Are fewer Actions better?}

In the experiment using Action 2 (row 6), we modify the action space to cover the same breadth of actions but have lesser number of actions. So the range remains the same but the step size between the actions increases. So instead of the actions being (0.0, 0.1, .. , 2.2), we now have (0.0, 0.2, .. , 2.2). We see that this experiment Pareto dominates the baseline i.e. it exceeds the baseline in both sparsity and accuracy. This is likely because with less number of actions to try, the agent is able to sample each action more and gain better knowledge of each action vis-a-vis the resultant performance metrics. Hence, it picks out better actions i.e. learns a better pruning policy given a particular layer in the network.   

\subsection{Generalization across ImageNet}
\label{imagenet}

To evaluate the performance of our agent on large scale tasks we experiment with the ImageNet dataset. We prune a pretrained ResNet-50 model using an iterative pruning scheme as mentioned in \cite{han2015learning} to preserve accuracy by providing gradual pruning targets for the network. We compare our performance with the state-of-the-art pruning algorithms AMC: AutoML for Model Compression \cite{he2018amc} and State of Sparsity \cite{gale2019state}. We prune more than 80\% and achieve comparable accuracy to state-of-the-art methods (see Table \ref{tab:resnet50} for full results).

\begin{table}
\centering
\small
\begin{tabular}{p{3cm}p{1.7cm}p{2cm}p{2cm}p{2cm}p{1.8cm}}
\toprule
\multirow{2}{*}{Method} & \multicolumn{4}{c}{ResNet-50} \\ \cmidrule(lr){2-6}
& Sparsity & Starting Acc. & Pruned Acc. & RL Episodes & Fine-tuning Epochs\\
\midrule
State of Sparsity & 80\% & 76.69\% & 76.52\% & NA & 153\\
AMC & 80\% & 76.13\% & 76.11\% & 400 & 120\\
PuRL & 80.27\% & 76.13\% & 75.37\% & 55 & 120\\
\bottomrule
\end{tabular}
\caption{We compare PuRL against the global state-of-the-art pruning results, not just for RL but for all pruning algorithms, and report the Top-1 accuracy performance on ImageNet. PuRL uses 85\% less RL episodes than AMC.}
\label{tab:resnet50}
\end{table}
Furthermore, PuRL finishes each RL training cycle in just 55 episodes, compared to 400 episodes required by AMC, due to the dense reward training procedure. We also conduct experiments on other state-of-the-art efficient architectures like MobileNet-V2 \cite{Sandler2018MobileNetV2IR} and EfficientNet-B2 \cite{tan2019efficientnet}. Referring to supplementary document, PuRL achieves more than 1.5x sparsity compared to AMC without much loss in accuracy. At the same time, PuRL achieves this performance on MobileNet-V2 without any changes in the underlying hyper-parameters compared to ResNet-50. Thus, PuRL can be easily used across architectures without the requirement of modifying the underlying MDP.

\section{Conclusion}
In this paper, we present PuRL - a fully autonomous RL algorithm for doing large scale compression of neural networks. By improving the rewards structure compared to current approaches, we shorten the training cycle of the RL agent from 400 to 55 episodes. We further do a detailed set of ablation experiments to determine the impact of each MDP component to the final sparsity and accuracy achieved by the agent. We achieve  results  comparable to current state-of-the-art pruning algorithms on the ImageNet dataset, sparsifying a ResNet-50 model by more than 80\% and achieving a Top-1 accuracy of 75.37\%. We also benchmark PuRL on other architectures like WideResNet-28-10 including already efficient architectures like MobileNet-V2 and EfficientNet-B2. Lastly, our algorithm is simple to adapt to different neural network architectures and can be used for pruning without a search for each MDP component.

% Acknowledgements should go at the end, before references and appendices

\acks{This research is supported by the Agency for Science, Technology and Research (A*STAR) under its AME Programmatic Funds (Project No.A1892b0026 and No.A19E3b0099).}

\vskip 0.2in
\bibliography{sample}

\begin{thebibliography}{15}
\providecommand{\natexlab}[1]{#1}
\providecommand{\url}[1]{\texttt{#1}}
\expandafter\ifx\csname urlstyle\endcsname\relax
  \providecommand{\doi}[1]{doi: #1}\else
  \providecommand{\doi}{doi: \begingroup \urlstyle{rm}\Url}\fi

\bibitem[Cheng et~al.(2017)Cheng, Wang, Zhou, and Zhang]{cheng2017survey}
Yu~Cheng, Duo Wang, Pan Zhou, and Tao Zhang.
\newblock A survey of model compression and acceleration for deep neural
  networks, 2017.

\bibitem[Deng et~al.(2009)Deng, Dong, Socher, Li, Li, and
  Fei-Fei]{imagenet_cvpr09}
J.~Deng, W.~Dong, R.~Socher, L.-J. Li, K.~Li, and L.~Fei-Fei.
\newblock {ImageNet: A Large-Scale Hierarchical Image Database}.
\newblock In \emph{CVPR09}, 2009.

\bibitem[Frankle et~al.(2019)Frankle, Dziugaite, Roy, and
  Carbin]{frankle2019stabilizing}
Jonathan Frankle, Gintare~Karolina Dziugaite, Daniel~M. Roy, and Michael
  Carbin.
\newblock Stabilizing the lottery ticket hypothesis, 2019.

\bibitem[Gale et~al.(2019)Gale, Elsen, and Hooker]{gale2019state}
Trevor Gale, Erich Elsen, and Sara Hooker.
\newblock The state of sparsity in deep neural networks.
\newblock \emph{arXiv preprint arXiv:1902.09574}, 2019.

\bibitem[Han et~al.(2015)Han, Pool, Tran, and Dally]{han2015learning}
Song Han, Jeff Pool, John Tran, and William Dally.
\newblock Learning both weights and connections for efficient neural network.
\newblock In \emph{Advances in neural information processing systems}, pages
  1135--1143, 2015.

\bibitem[Han et~al.(2016)Han, Mao, and Dally]{Han2015DeepCC}
Song Han, Huizi Mao, and William~J. Dally.
\newblock Deep compression: Compressing deep neural network with pruning,
  trained quantization and huffman coding.
\newblock \emph{International Conference on Learning Representations}, 2016.

\bibitem[He et~al.(2018)He, Lin, Liu, Wang, Li, and Han]{he2018amc}
Yihui He, Ji~Lin, Zhijian Liu, Hanrui Wang, Li-Jia Li, and Song Han.
\newblock Amc: Automl for model compression and acceleration on mobile devices.
\newblock In \emph{Proceedings of the European Conference on Computer Vision
  (ECCV)}, pages 784--800, 2018.

\bibitem[Krizhevsky et~al.()Krizhevsky, Nair, and Hinton]{CIFAR100}
Alex Krizhevsky, Vinod Nair, and Geoffrey Hinton.
\newblock Cifar-100 (canadian institute for advanced research).
\newblock URL \url{http://www.cs.toronto.edu/~kriz/cifar.html}.

\bibitem[Lin et~al.(2017)Lin, Rao, Lu, and Zhou]{NIPS2017_6813}
Ji~Lin, Yongming Rao, Jiwen Lu, and Jie Zhou.
\newblock Runtime neural pruning.
\newblock In I.~Guyon, U.~V. Luxburg, S.~Bengio, H.~Wallach, R.~Fergus,
  S.~Vishwanathan, and R.~Garnett, editors, \emph{Advances in Neural
  Information Processing Systems 30}, pages 2181--2191. Curran Associates,
  Inc., 2017.
\newblock URL
  \url{http://papers.nips.cc/paper/6813-runtime-neural-pruning.pdf}.

\bibitem[Mnih et~al.(2013)Mnih, Kavukcuoglu, Silver, Graves, Antonoglou,
  Wierstra, and Riedmiller]{mnih2013playing}
Volodymyr Mnih, Koray Kavukcuoglu, David Silver, Alex Graves, Ioannis
  Antonoglou, Daan Wierstra, and Martin Riedmiller.
\newblock Playing atari with deep reinforcement learning.
\newblock \emph{NIPS Deep Learning Workshop}, 2013.

\bibitem[Ng et~al.(1999)Ng, Harada, and Russell]{ng1999policy}
Andrew~Y Ng, Daishi Harada, and Stuart Russell.
\newblock Policy invariance under reward transformations: Theory and
  application to reward shaping.
\newblock In \emph{ICML}, volume~99, pages 278--287, 1999.

\bibitem[Sandler et~al.(2018)Sandler, Howard, Zhu, Zhmoginov, and
  Chen]{Sandler2018MobileNetV2IR}
Mark Sandler, Andrew~G. Howard, Menglong Zhu, Andrey Zhmoginov, and Liang-Chieh
  Chen.
\newblock Mobilenetv2: Inverted residuals and linear bottlenecks.
\newblock \emph{2018 IEEE/CVF Conference on Computer Vision and Pattern
  Recognition}, pages 4510--4520, 2018.

\bibitem[Strubell et~al.(2019)Strubell, Ganesh, and
  McCallum]{strubell2019energy}
Emma Strubell, Ananya Ganesh, and Andrew McCallum.
\newblock Energy and policy considerations for deep learning in nlp.
\newblock \emph{57th Annual Meeting of the Association for Computational
  Linguistics (ACL)}, 2019.

\bibitem[Tan and Le(2019)]{tan2019efficientnet}
Mingxing Tan and Quoc~V Le.
\newblock Efficientnet: Rethinking model scaling for convolutional neural
  networks.
\newblock \emph{Proceedings of the 36th International Conference on Machine
  Learning}, 2019.

\bibitem[Zagoruyko and Komodakis(2016)]{BMVC2016_87}
Sergey Zagoruyko and Nikos Komodakis.
\newblock Wide residual networks.
\newblock In Edwin R.~Hancock Richard C.~Wilson and William A.~P. Smith,
  editors, \emph{Proceedings of the British Machine Vision Conference (BMVC)},
  pages 87.1--87.12. BMVA Press, September 2016.
\newblock ISBN 1-901725-59-6.
\newblock \doi{10.5244/C.30.87}.
\newblock URL \url{https://dx.doi.org/10.5244/C.30.87}.

\end{thebibliography}

% Appendix goes to a new page

\newpage

\appendix

\section{PuRL Algorithm}

Algorithm \ref{alg:auto_prune} describes a DQN procedure which learns to select the sparsity threshold for each layer in the model. At the start of each episode, the original model with pre-trained weights is loaded. The agent then takes an action and the layer is pruned by that amount. The model is then retrained for one epoch on a small subset of training data (1000 images out of the 1.2million images in ImageNet). After that, the validation accuracy is calculated, to effect the state transition and the reward is then calculated using the validation accuracy and the pruning percentage. The training is done for \textit{max\_episodes} episodes. Once, training is completed, the model is pruned using the trained agent and then fine-tuned on the full ImageNet dataset.

\begin{algorithm}[h]
\caption{The PuRL Algorithm}
\label{alg:auto_prune}
\begin{algorithmic}[1]

\State \textbf{Stage 1: Train DQN agent}
\State \textit{episodes} $\gets$ 0
\While {\textit{episodes} $\leq$ \textit{max\_episodes}}
\State \textit{model} $\gets$ load original model
\For {each layer \textit{t} in the \textit{model}}
\State $a_t$ $\gets$ Sample action from DQN agent
\State Prune layer \textit{t} using $a_t$ 
\State Retrain \textit{model} on small subset of data for 1 epoch
\State Calculate reward $r_{t}$ and new state $s_{t}$ based on resultant sparsity and accuracy achieved
\State Return $r_{t}$ and $s_{t}$ to agent
\EndFor
\State \textit{episodes} $\gets$ \textit{episodes} + 1
\EndWhile\\

\State \textbf{Stage 2: Prune and fine-tune model}
\State \textit{model} $\gets$ load original model
\State Prune \textit{model} using the trained DQN agent (averaged over 5 episodes)
\State Fine-tune \textit{model}
\State \textbf{return} \textit{model}
\end{algorithmic}
\end{algorithm}

\section{Experimental Results}

\subsection{Understanding the RL Design Space}
\label{ablation_app}

\subsubsection{Is Absolute Magnitude better than Standard Deviation?}

Referring to the Magnitude Target based ablation Table \ref{tab:ablation_results} (row 3), we compare absolute magnitude based pruning to standard deviation based pruning (Section \ref{Approach}). For absolute magnitude, we set a sparsity target for a layer and then remove all small weights until we hit the desired sparsity level. As we observe, both the sparsity and accuracy results are lower in this case as compared to our baseline experiment (dense rewards). Thus, standard deviation based pruning is better. 

\subsubsection{Does Reward Shaping help?}

In the experiments using Reward 2 and Reward 3 (Table \ref{tab:ablation_results}), we investigate if reward shaping can help the agent achieve higher accuracy. For Reward 2, we allow the agent to receive positive rewards if it surpasses the given accuracy target (Equation \ref{eq:RewEq1}). This is in contrast to the baseline reward function in which there is a cap on the maximum reward that the agent can achieve i.e. zero.  
\begin{equation}
\textrm{R}_2 = - S ((A/T_{A} -1) + \max (1 - PP/T_{PP} , 0)) 
\label{eq:RewEq1}
\end{equation}
For Reward 3, we build on Reward 2 and give a cubic reward to the agent. The agent now sees cubic growth in positive reinforcement as it approaches and surpasses the accuracy target (Equation \ref{eq:RewEq2}). Hence, by taking the same step size towards accuracy improvement as compared to Reward 2, the agent now gets rewarded more for it.  
\begin{equation}
\textrm{R}_3 = - S ( ((A/T_{A})^3 -1) + \max (1 - PP/T_{PP} , 0))    
\label{eq:RewEq2}
\end{equation}
The performance of both these functions is close to the baseline (dense rewards), but the baseline still outperforms them. The added complexity of these functions might require the agent to sample more steps to learn them well. Hence, given a tight training budget, the baseline reward function performs well.

\begin{table*}[t]
\small
\begin{tabular}{m{32mm}m{6mm}m{8mm}m{9mm}m{9mm}m{9mm}m{9mm}m{16mm}m{16mm}}
\toprule
\multirow{2}{*} {Experiment} &  \multirow{2}{*} {State} &  \multirow{2}{*} {Action} & \multicolumn{4}{c} {Reward Space} &  \multirow{2}{*}{Accuracy} &  \multirow{2}{*}{Sparsity}  \\\cmidrule(lr){4-7}
&Space&Size& \centering Prune Penalty& \centering Acc. Penalty& \centering Acc. Upside&\centering Cubic Upside&& \\%\hline
\midrule
Low Dimensional State & \centering3 & \centering0.2 & \centering\checkmark & \centering\checkmark  & & & 47.6 $\pm$ 2.3 & 87.4 $\pm$ 1.3  \\ %\hline
Higher Dimensional State & \centering108 & \centering0.2 & \centering\checkmark & \centering\checkmark  & & &  51.0 $\pm$ 0.3 & 82.0 $\pm$ 2.0  \\ %\hline
\bottomrule
\end{tabular}
\caption{Follow-up experiment on perturbing the State space on the ImageNet dataset. Error denotes standard error as measured on 3 trials. The higher dimensional state space (108 dimensions) performs better than the simple low dimensional state space (3 dimensions)}
\label{tab:ablation_results2}
\end{table*}

\subsubsection{Is more information better for the agent?}

In the last experiment with State 2 (Table \ref{tab:ablation_results}), we vary the state space and make it 108 dimensional instead of 3 dimensional. The idea here is to give the agent more information on the state representation (See Section \ref{Approach} for details). We see that in this experiment, the agent achieves less accuracy than the baseline however, prunes more than it. Hence, none of the experiments Pareto dominate each other and its inconclusive to determine which one is better. To get more evidence on this, we carry out a further ablation on the ImageNet dataset. Referring to Table \ref{tab:ablation_results2}, we see that the 108 dimensional state outperforms the 3 dimensional state. Hence, more information is indeed better and we use this feature in the final configuration. 

\subsection{Scaling PuRL to CIFAR100}
\label{cifar100}

We first experiment with PuRL on the WideResNet-28-10 architecture \cite{BMVC2016_87} on the CIFAR-100 \cite{CIFAR100} dataset. We compare it to a uniform pruning baseline where every layer is pruned by the same amount to achieve a target sparsity of 93.5\%. PuRL outperforms the baseline in Table \ref{tab:cifar100} on both the sparsity and final accuracy.

\begin{table}
\small
\centering
\begin{tabular}{p{2cm}p{1.5cm}p{2cm}p{2cm}}
\toprule
\multirow{2}{*}{Method} & \multicolumn{3}{c}{WideResNet-28-10} \\ \cmidrule(lr){2-4}
& Sparsity & Top-1 Acc. Pre-Pruning & Top-1 Acc. Post-Pruning\\ 
\midrule
Baseline & 93.50\% & 82.63\% & 72.42\%\\
PuRL & 93.90\% & 82.63\% & 80.63\%\\
\bottomrule
\end{tabular}
\caption{Comparison of the PuRL algorithm to a uniform pruning baseline on the WideResNet-28-10 architecture on CIFAR-100 dataset. PuRL beats the baseline by a huge margin}
\label{tab:cifar100}
\end{table}

\begin{table}
\small
\centering
\begin{tabular}{p{2cm}p{1.5cm}p{1.5cm}p{2cm}p{2cm}}
\toprule
\multirow{2}{*}{Method} & \multicolumn{4}{c}{MobileNet-V2} \\ \cmidrule(lr){2-5}
& Sparsity &  Flops reduction & Top-1 Acc. Pre Pruning & Top-1 Acc. Post Pruning\\
\midrule
AMC & Not reported & 30\% & 71.8\% & 70.8\% \\ 
PuRL & 43.3\% & 47.9\% & 71.9\% & 69.8\% \\
\bottomrule
\end{tabular}
\caption{Comparison of PuRL to AMC for the MobileNet-V2 architecture.}
\label{tab:expt_models2}
\end{table}

\begin{table}[h!]
\small
\centering
\begin{tabular}{p{2cm}p{1.5cm}p{2cm}p{2cm}}
\toprule
\multirow{2}{*}{Method} & \multicolumn{3}{c}{EfficientNet-B2} \\ \cmidrule(lr){2-4}
& Sparsity & Top-1 Acc. Pre Pruning & Top-1 Acc. Post Pruning \\
\midrule
Baseline & 59.0\% & 79.8\% & 68.9\% \\ 
PuRL & 59.5\% & 79.8\% & 74.5\% \\ 
\bottomrule
\end{tabular}
\caption{Comparison of PuRL to uniform pruning baseline on the state-of-the-art EfficientNet-B2 architecture on the ImageNet dataset. PuRL outperforms the baseline on both the sparsity and accuracy}
\label{tab:expt_models3}
\end{table}

\subsection{Generalization across ImageNet}
\label{imagenet_app}

We also conduct experiments on other state-of-the-art efficient architectures on ImageNet to see whether our pruning algorithm can make these architectures even more sparse. We experiment on MobileNet-V2 \cite{Sandler2018MobileNetV2IR} and EfficientNet-B2 \cite{tan2019efficientnet}. Referring to Table \ref{tab:expt_models2}, PuRL achieves more than 1.5x sparsity compared to AMC without much loss in accuracy. 

At the same time, PuRL achieves this performance on MobileNet-V2 without any changes in the underlying hyper-parameters compared to ResNet-50. Thus, PuRL can be easily used across architectures without the requirement of modifying the underlying MDP.  For EfficientNet-B2, Table \ref{tab:expt_models3}, we compare PuRL to a uniform pruning baseline. PuRL outperforms the baseline on both sparsity and final accuracy, achieving an accuracy improvement of more than 5\%. In this case as well, we set the exact same hyper-parameters and MDP setting as that of ResNet-50 and MobileNet-V2. However, since Efficient-B2 is very deep, having 116 layers compared to 54 in ResNet-50, we do early-stopping of the RL episode, to make the training even faster. We stop the episode if the test accuracy drops to less than 0.1\% and move on to the next episode.

\end{document}